\documentclass[letterpaper, 10 pt, conference]{ieeeconf}  % Comment this line out if you need a4paper

\IEEEoverridecommandlockouts                              % This command is only needed if 
\usepackage{graphics} % for pdf, bitmapped graphics files
\usepackage{epsfig} % for postscript graphics files
\usepackage{mathptmx} % assumes new font selection scheme installed
\usepackage{times} % assumes new font selection scheme installed
\usepackage{amsmath} % assumes amsmath package installed
\usepackage{amssymb}  % assumes amsmath package installed
\usepackage{graphicx}
\usepackage{cite}
\usepackage{float}
\usepackage{wrapfig}
\usepackage{sidecap}
\usepackage{footmisc}
\usepackage{booktabs}
\usepackage{caption}
\title{\LARGE \bf
Estimating Pedestrian Crossing States Based on Single 2D Body Pose
}

\author{Zixing Wang$^{1}$ and Nikolaos Papanikolopoulos$^{2}$% <-this % stops a space
\thanks{$^{1}$Zixing Wang is with Department of Computer Science and Engineering,
        University of Minnesota, Minneapolis, MN 55414, USA.
        {\tt\small wang7923@umn.edu}}%
\thanks{$^{2}$Nikolaos Papanikolopoulos is with Faculty of Department of Computer Science and Engineering,
        University of Minnesota, Minneapolis, MN 55414, USA.
        {\tt\small npapas@cs.umn.edu}}%
}

\begin{document}

\maketitle
\thispagestyle{empty}
\pagestyle{empty}

%%%%%%%%%%%%%%%%%%%%%%%%%%%%%%%%%%%%%%%%%%%%%%%%%%%%%%%%%%%%%%%%%%%%%%%%%%%%%%%%
\begin{abstract}

The Crossing or Not-Crossing (C/NC) problem is important to autonomous vehicles (AVs) for safe vehicle/pedestrian interactions. However, this problem setup often ignores pedestrians walking along the direction of the vehicles' movement (LONG). To enhance the AVs' awareness of pedestrians behavior, we make the first step towards extending the C/NC to the C/NC/LONG problem and recognize them based on single body pose. In contrast, previous C/NC state classifiers depend on multiple poses or contextual information. Our proposed shallow neural network classifier aims to recognize these three states swiftly. We tested it on the JAAD dataset and reported an average 81.23\% accuracy. Furthermore, this model can be integrated with different sensors and algorithms that provide 2D pedestrian body pose so that it is able to function across multiple light and weather conditions.

\end{abstract}

%===============================================================================

\section{Introduction}
\label{sec:Introduction}
In 2017, 5,977 pedestrian deaths in USA were reported by the United States National Highway Traffic Safety Administration (NHTSA). Among these fatalities, 5,890 pedestrians were killed by single or multiple motor vehicles. When the accidents happened, 84.4\% (4,529) of victims were struck by the front of the vehicles~\cite{c3}. According to these data, we believe that an approach to avoid hitting pedestrians by the front of vehicles, which accounts for the highest proportion of fatal vehicle-human accidents, is critical for AVs.

Simply obeying traffic rules is not enough for AVs to avoid vehicle-human accidents. Two mainstream classes of methods to remedy this problem include prediction-based and estimation-based methods. However, these methods have some flaws, which may impact their potential for deployment. 

Previous prediction-based research regress pedestrians' incoming moving trajectory or predict their crossing intention based on previous trajectories~\cite{c4} or velocities~\cite{c5}. More recent research takes contextual elements~\cite{c2} such as weather conditions, time in the day, etc., into consideration to improve performance. However, predicting pedestrians' intention remains a challenging problem due to their arbitrary upcoming motion~\cite{c8}. They can decide to change moving direction, stop crossing, etc., within a second. Moreover, they are influenced by multiple internal and external factors. For instance, arbitrary actions such as failure to yield right of way, improper crossing of the roadway or the intersection, darting or running into the road and failure to obey traffic signs, signals, or officer commands led to 1,788 (29.9\%), 1,268 (21.2\%), 592 (9.9\%), and 266 (4.5\%) fatalities~\cite{c3} in 2017. Thus, it is very difficult to predict pedestrian intention based on prior data (trajectory, velocity, etc.) or contextual information, which is unquantifiable and hard to identify, within a short time window. In addition, most of these approaches need significant computational power and work offline, which limit their deployment on AVs as a real-time safety mechanism.

\begin{figure}[t]
    \centering
    \includegraphics[width=0.48\textwidth]{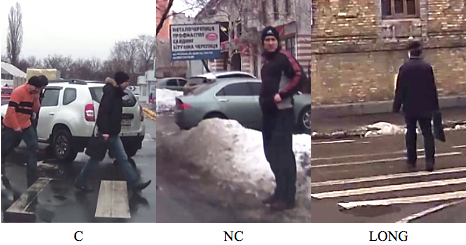}
    \caption{Our focus: recognize crossing (C), not-crossing (NC) and walking along the direction of vehicles' movement (LONG) based on single body pose within a short time. Note that previous related works categorize both not-crossing and parallel walking as NC.}
    \label{fig:diff_pose}
\end{figure}

\begin{figure*}[t]
    \centering
    \includegraphics[width=\textwidth]{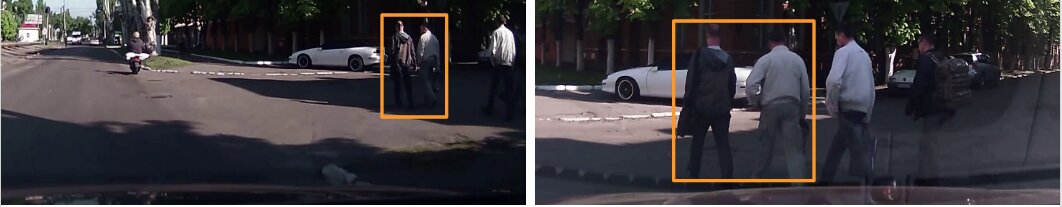}
    \caption{The pedestrians inside orange bounding boxes are classified as NC and C in the left and right figures. }
    \label{fig:lat}
\end{figure*}

Earlier estimation-based research~\cite{c1,c10,c11} simply classifies pedestrians' action to cross and not-cross (C/NC). Crossing indicates pedestrians' movement that is lateral to vehicle, and the rest belongs to not-crossing. This is a dangerous classification, especially when the vehicle turns at an intersection. For instance, as Fig.~\ref{fig:lat} shows, before the vehicle turns right, two pedestrians bounded by an orange box will be classified as NC. However, in the middle of turning, they would be classified in a C state by using the same criteria. This ambiguity can lead to severe traffic accidents. (We extend the categories from C/NC to C/NC/LONG in this work, which allows AVs to sense pedestrian behavior in both parallel and lateral directions to avoid this dilemma.) In addition, most of these works make the estimation based on the appearance of pedestrians. So, in low-light or severe weather conditions, the performance will drastically deteriorate. 

We believe that the state estimation is more promising and applicable than intention prediction. Our reasons are the following. Firstly, when vehicles move on roads, information about the surroundings including pedestrians updates very frequently and includes a lot of recent content. State estimation can capture this varying information and can guide the decision making more effectively. Moreover, intention prediction is using a lot of subtle cues that often could not capture
the real actions or intentions of pedestrians. Finally, intention is often inferred from limited in time information that may not be representative of what the actual state of the pedestrian movement is.

To make AVs interact with pedestrians more safely and address the problems of the estimation type research (stated before), we propose a neural network classifier performing the C/NC/LONG task based on a single 2D body pose. The idea is illustrated in Fig.~\ref{fig:diff_pose}. Since our network works in AVs and is designed to avoid collisions, it is very sensitive to running speed and computational resources. In fact, our 2D pose contains 36 floating point numbers. Due to these two reasons, the network is very shallow and has a limited number of parameters. By extending the previous approach to C/NC/LONG problems, the AVs will have better awareness of pedestrians' moving/crossing states. Moreover, dilemmas such as what is illustrated in Fig.~\ref{fig:lat} could be addressed in a more comprehensive way. Moreover, we replace the pedestrians' appearance by the 2D body pose since we believe it is a strong indicator of pedestrians motion state and current state-of-the-art algorithms and sensors integrated on AVs are able to provide reliable pedestrian detection, tracking, and pose estimation~\cite{c12,c13,c14,c15,c16,c17,c18} data in various weather and light conditions to enable all-time onboard operation. We believe our work can provide valuable reference to AVs to avoid Human-Vehicle collisions. For instance, pose estimators provide tracking (e.g., Bounding Boxes (BB)) and 2D pose data. If a BB is near or at the sidewalk with a C tag, then the AV should stop to avoid potential collisions. But if the label is NC or LONG, the AV can start or keep moving, which will not affect traffic efficiency. In contrast, if a BB is at the roadways, then no matter what label is associated with it, the AV should immediately stop.

The philosophy behind our approach is: we place priority on the well-being of humans and thus, AVs should always yield the road right once crossing pedestrians are detected in front of the vehicle. A neural network that is able to robustly output accurate estimation of a crossing state within very short period of time would allow AVs to have enough time to perform smoother and safer maneuvers to prevent collisions. This has an impact on all stakeholders since it makes passengers more comfortable and pedestrians more confident to cross~\cite{c1}. In addition, even if a vehicle-pedestrian collision is not avoidable, the proposed approach will be able to reduce the chance of causing hospitalization-required injuries~\cite{c9,c1}.

\section{Related Work}
\label{sec:Related Work}
In this section, we mainly review prior research in the area of predicting pedestrian behavior based on body pose and C/NC classification. 

The research in \cite{c19,c20} leverages the contour of pedestrians to predict their intentions. Moreover, posture~\cite{c21,c22} and body language~\cite{c23,c24} were also studied for the purpose of predicting pedestrian intentions. The research in~\cite{c24,c25,c26} tends to approximate head and body orientation to estimate pedestrian intentions. However, in~\cite{c26} the experiments showed that head detection did not provide useful data for the C/NC task. The work in~\cite{c27} combined lateral speed, orientation, pose, and abstract scene information to feed them to a neural network, which was able to predict impending actions. 

Most studies mentioned above focus on partial features of the body pose. The work in~\cite{c28} suggested that without information about the pedestrians' posture and body motion, the detection of pedestrians' intention change will be delayed. The baseline evaluation of the JAAD dataset~\cite{c2} supported this conclusion. They compared C/NC task performance between approaches with full body features and appearance. They specifically focused on sub-appearance of partial feature sets. The results suggested that the latter would not help to improve C/NC task performance. In~\cite{c29}, instead of appearance, they accumulate and update the full body pose (i.e., skeleton key-points) and features over time by a sliding time-window as input to their support vector machine (SVM) classifier. They obtained best C/NC task performance on the Daimler's dataset~\cite{c30}. Later they improved their method and tested the approach on the JAAD dataset. 
\raggedbottom

\section{Method}
\label{sec:Method}
In this section, we introduce the way to pre-process data in \ref{subsec:dp} and the structure of our neural network in \ref{subsec:clsf}.

\subsection{Data Processing}

\label{subsec:dp}
\begin{figure}[t]
\centering
\includegraphics[width=0.48\textwidth]{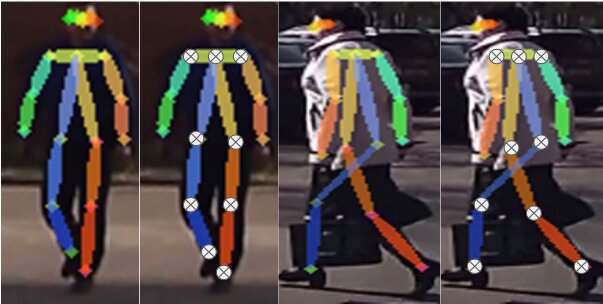}
\caption{9 most stable key-points.}
\label{fig:star}
\end{figure}

The typical COCO format body pose has 18 key-points. However, when pedestrians cross in front of the vehicle, one of their arms is likely partially or fully invisible to the ego-camera on the vehicle, so as key-points on the face. To resolve this issue, we follow a similar pose preprocessing procedure to the work described in ~\cite{c1}. A pruned pose only contains the 9 most stable key-points\footnote{Right \& left shoulder, neck, right \& left hip, right \& left knee, right \& left ankle.}, which represent shoulders and legs (Fig.~\ref{fig:star} shows relative positions of these points on the human body). These key-points indicate essential action information including motion state (start walking/ keep walking/ stop walking/ stand) and movement orientation. To eliminate the negative influence caused by the different pose scales, we translate and normalize them. First, we set the pose center at the neck (assign (0, 0) to the neck key point). Then, for each pose we respectively compute the normalization factor \textsl{n}: $n = max(pose.y) - min(pose.y)$ and normalize the x, y coordinates.
The \textsl{pose.x} represents all x coordinates while the \textsl{pose.y} represents all the y coordinates associated with the pose. In Fang's work~\cite{c1}, they computed 396 features including distances, angle, etc. for their Support Vector Machine (SVM) classifier. In contrast, we have the neural network to extract and interpret features.

\subsection{Neural Network}
\label{subsec:clsf}
\begin{figure}[b]
    \centering
    \includegraphics[width=0.48\textwidth]{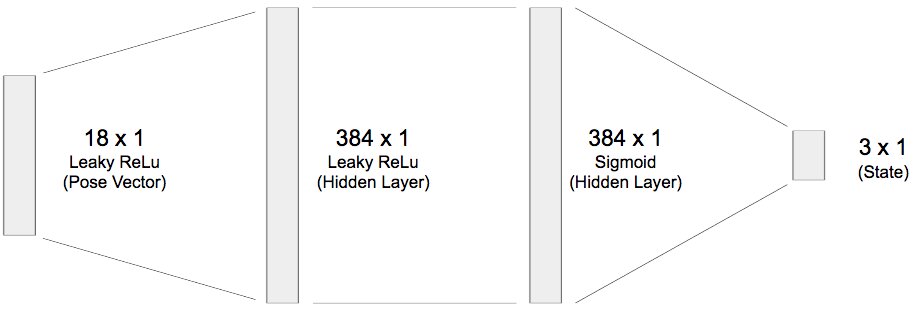}
    \caption{Structure of the C/NC/LONG Classifier.}
    \label{fig:nn}
\end{figure}

As we mentioned in the introduction section, the real-time C/NC/LONG classification task is extremely sensitive to time constraints and mobile platforms generally have limited computational power. So, we designed a shallow neural network, which only contains 2 hidden fully connected layers and requires limited resources. The network takes the x and y coordinates of the body pose as input data and outputs the probability for each state. Considering the very limited size of the input data (18 floating point numbers), we have trust in the ability of extracting and understanding features of our neural network despite the limited number of parameters. The details of our network are illustrated in Fig.~\ref{fig:nn}.

\section{Experiments}
\label{sec:experiment}
In order to evaluate the performance of our classifier and create a pose annotation of the JAAD dataset, we use the AlphaPose\cite{c17} as pose estimator, which is briefly introduced in \ref{sec:pose}. Moreover, we describe the overall organization of our dataset and training procedure in Sections \ref{sec:dataset} and \ref{sec:training}. Then, the general performance including numeric information and sample images is reported in Section \ref{sec:classifier}.

\begin{figure*}[t]
    \centering
    \includegraphics[width=\textwidth]{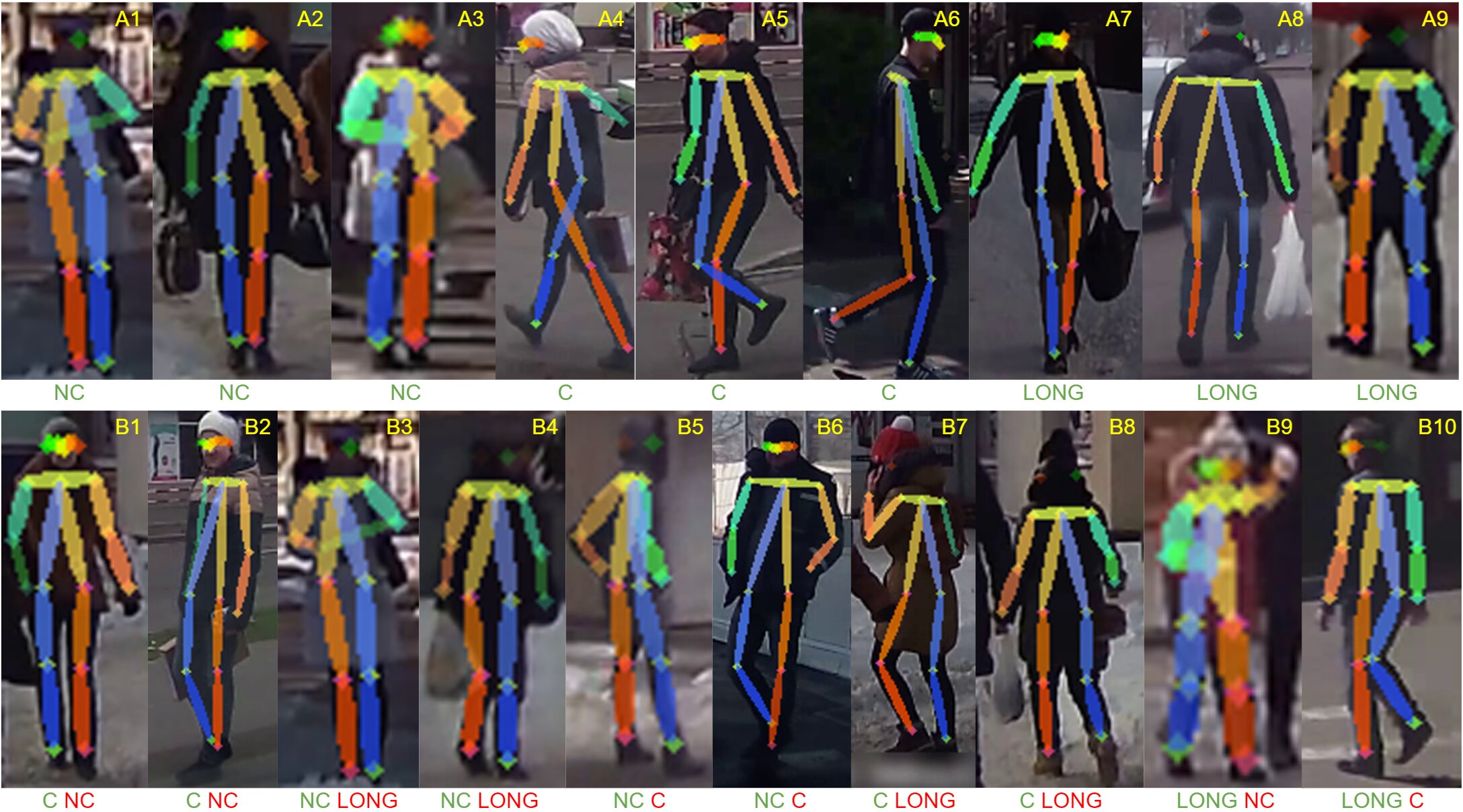}
    \caption{The first and second row show success and failure cases. Green texts indicate the ground truth for each image and red texts are erroneous predictions. If there is only green text under a image, it indicates a correct prediction.}
    \label{fig:smp} 
\end{figure*}

\subsection{Pose Estimation}
\label{sec:pose}

AlphaPose is an advanced real-time pose estimator which firstly achieved 70+ mAP (72.3 mAP) on the COCO dataset and 80+ mAP (82.1 mAP) on the MPII dataset. We use it to generate the COCO model body poses. Then we combine qualified poses, which are selected by the human, with manual annotated poses as training data. It is worth noting that we run AlphaPose without the Poseflow~\cite{c31} function, which matches poses that correspond to the same person across frames, since the JAAD dataset provides the ground truth Bounding Box (BB) for each subject across frames. We run the Windows version PyTorch implementation of AlphaPose with parameters: --conf 0.05, --nms 0.8, which are the confidence thresholds for human detection.
\raggedbottom

\subsection{Dataset}
\label{sec:dataset}
The JAAD (Joint Attention for Autonomous Driving)~\cite{c2} is a dataset that provides data including but not limited to BBs. It includes Behavior Tags of subjects on each frame and the moving direction of pedestrians on each video acquired in real-world driving conditions. Due to its large size, approximately 2.2k unique pedestrians and 337k bounding boxes, our pose annotations only cover part of the whole dataset. 

AlphaPose generates poses for qualified images (width is greater than 60 pixels) cropped from the original JAAD frames. Before manual filtering, we remove poses that have average confidence score lower than 0.6 and with key-points' scores lower than 0.5. Note that the average confidence score only takes 9 body key-points (Fig.~\ref{fig:star}) into account. In the end, after inspection of them, we have 12,756 manual annotated and auto-generated poses in total. Furthermore, since the JAAD does not directly provide labels of C, NC, and LONG, we map the behavioral label to them as follows. We term as C the behavioral labels of walking, crossing, moving fast, moving slow, slow down, speed up, and clear path when the corresponding subject possesses an LAT label, which means the pedestrian is crossing in front of the car. In contrast, if the subject is labeled as LONG in JAAD, which indicates the same behavior as our LONG label does, we term it to this label set. And, stopped and standing labels belong to NC. After mapping, we have 4,805 C, 4,096 LONG, and 3,855 NC labels in our pose dataset.

We take the first 10,544 (84\%) images for training and the rest (16\%) for testing. The dataset was split to 84:16 since in this way the amount of each class in the training and testing sets is relatively balanced. In the testing set, there are 626 (28.3\%) C, 862 (38.9\%) NC, and 724 (32.7\%) LONG instances. Furthermore, in the training set there are 4,179 (39.6\%) C, 2,993 (28.3\%) NC, and 3,372 (31.9\%) LONG instances.

\subsection{Training}
\label{sec:training}
We use PyTorch~\cite{c32} to build, train, and test our neural network with an NVIDIA GTX 1080 Ti. We use the stochastic gradient descent (SGD) optimizer with an initial learning rate of 2e-1 and a momentum of 0.5, a PyTorch ReduceLROnPlateau scheduler with a 0.5 decay rate and 3 patience and a batch size of 128. In addition, we use cross-entropy to evaluate loss. We trained our neural network in 50 epochs and it converged within 10 epochs. Note that our model needs a high initial learning rate to converge and the Adam optimizer is less likely to lead to convergence in our experience. In addition, the size of this neural network is 2,378KB as a python pickle file without computational graph, which is easy to be deployed.

\subsection{Classifier}
\label{sec:classifier}

\begin{table}[b]
\
\centering
\caption{
Performance of the Classifier on the Testing Set.}
\resizebox{0.48\textwidth}{!}{%
\footnotesize
\begin{tabular}{cccc}
\hline
\textbf{Class} & \textbf{Total} & \textbf{Correct} & \textbf{Accuracy} \\ 
\hline
C & 626 & 439 & 70.13\% \\
NC & 862 & 797 & 92.46\% \\
LONG & 724 & 561 & 77.48\% \\
ALL & 2212 & 1797 & 81.23\% \\ 
\hline
\end{tabular}%
}
\label{acc}
\end{table}

\begin{figure*}[h]
    \centering
    \includegraphics[width=\textwidth]{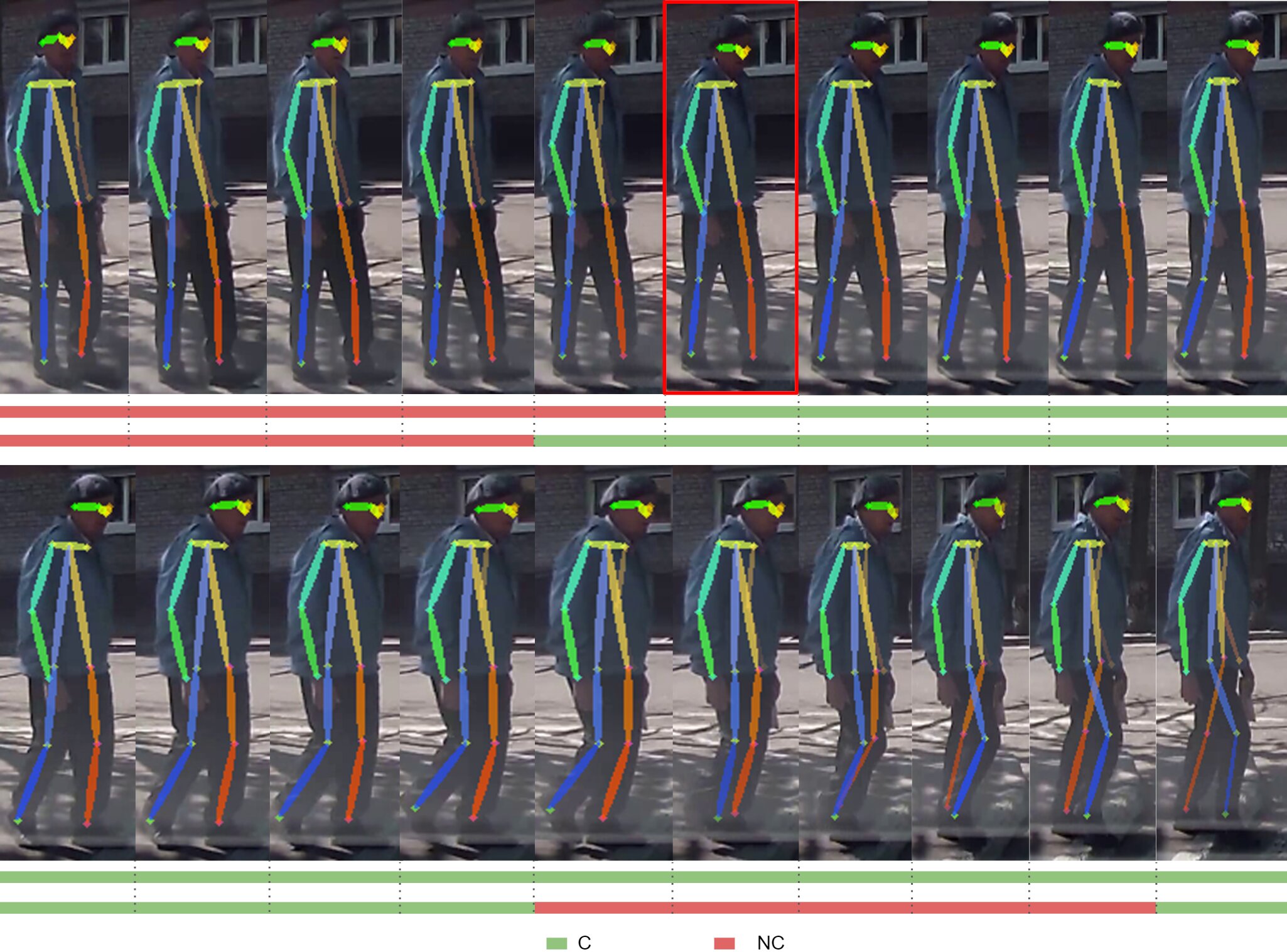}
    \caption{A typical TTE test sequence. The highlighted frame is the Time-To-Event (TTE) frame. The first and second line of the bar under each frame represent ground-truth and estimation. Bars in green and red represent the C and NC states.} 
    \label{fig:seq_pose}
\end{figure*} 

We make independent and sequential tests to evaluate the absolute and realistic performance of the classifier. However, to the best of our information and knowledge (at the moment) our classifier is the only work designed for the C/NC/LONG task. Although it is possible to pool the classes of interest to do the comparisons, we cannot simply pool C/NC sections from our C/NC/LONG results. This is due to the fact that the NC category of earlier works contains the LONG category of our work (previous NC = our NC + LONG) as Fig.~\ref{fig:diff_pose} indicates. So, even if we take all the LONG instances out of the NC category of previous works, they are still incomparable with our work since they are designed for totally different tasks. In addition, although we reported acceptable results, our neural network is a first step towards a new set of algorithms that are applicable to this domain and has much space for improvement (as expected). We believe it is a good baseline experiment in this field and a step for future investigations. Due to these reasons, in this section we focus on reporting results and analyzing potential reasons for the failure cases.

For the independent evaluation, Table ~\ref{acc} reports the overall and category accuracy and Fig.~\ref{fig:smp} shows samples of success and failure cases. We calculate the accuracy according to the widely used definition: $AC = C/ T * 100\%$, where AC, C, and T stands for accuracy rate, total correct prediction, and total poses tested, respectively. We achieved 81.23\% general accuracy rate.

According to Table ~\ref{acc}, we can see that the model has better performance on recognizing NC state poses. The reason could be that the features of standing pose are more stable and obvious than C and LONG state poses since the former are from static but the latter are from dynamic actions. Furthermore, through observation of the NC failure cases we notice that while standing, particularly close to curbside, some of the subjects would pace in-situ or move legs without exact purpose, such as in Fig.~\ref{fig:smp} B6, which makes their pose similar to C or LONG state. Moreover, people's unbalanced and ready-to-walk standing poses (Fig.~\ref{fig:smp} B5 can also lead to failed predictions). According to the error test results, we find that a large portion of failed predictions are incorrectly recognizing C or LONG poses as an NC state. The essential reason could be that the pose is similar to the NC state when the distance of legs is very close while walking as shown in Figs. \ref{fig:smp} (B1, B2, and B9). 

In addition, we report the Time-To-Event (TTE) test. It evaluates the classifier's response time to a state change of pedestrians, which partially reflects how early an AV starts maneuvering to avoid collision. To simplify our description, we introduce some related concepts and notation. The Time-To-Event frame (TTEF) is the first frame after the change of pedestrian's crossing state. For instance, in Fig.~\ref{fig:seq_pose}, the subject's state changes from NC to C. The first five frames are in NC state and the last fifteen are in C state. So the sixth frame is the TTE frame. The Response frame (RF) corresponds to the number of frames before (RF $<$ 0) or after (RF $>$ 0) the classifier starts responding to state changes. When the classifier responds to a state change and generates no less than nine correct estimations, the result is considered as a confident estimation (CE). Moreover, because there is no previous research report on the TTE test, we indirectly compare the response time of the classifier (CRT) with the human driver's response time (DRT). CRT is defined as $CRT = RF * 1/FrameRate$. The unit of CRT and DRT is the second (s).

\begin{figure}[t]
    \centering
    \includegraphics[width=0.48\textwidth]{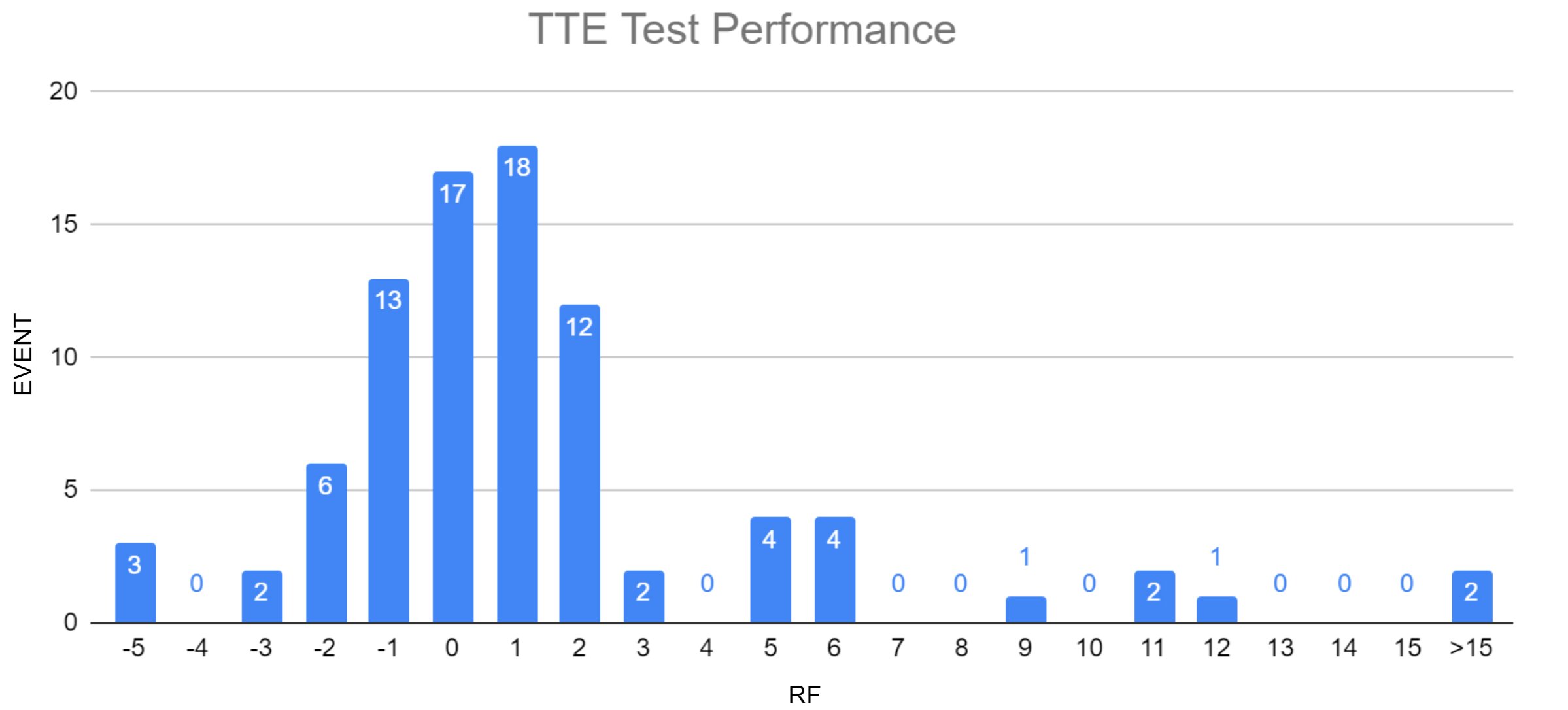}
    \caption{The x-axis is the sample space of RF, the height of each bar represents the number of occurrence of each RF. On the x-axis in this graph the unit is the frame.}
    \label{fig:tte}
\end{figure}

We apply the TTE test on 87 sequences, which are in chronological order. As Fig.~\ref{fig:seq_pose} shows, each sequence contains 20 frames that 5 and 15 frames of them are before and after the TTEF (TTEF belongs to the later state). The corresponding CRT space is from -1/6 to 1/2 s. Fig.~\ref{fig:tte} shows the general result. In total, the classifier successfully recognized state changes no later than 15 frames after the TTEF in 85 sequences. Another two late CEs happened in the 19 and 22 RF. The average and median RF (Events for which RF $>$ 15 are ignored) is 4.238 and 1, which is 0.141267 s and 0.03333 s in the CRT. As the chart shows, the classifier is quite sensitive to state changes and even the latest CE has a CRT of 0.3667 s. In 2018, 50 volunteers (25 female and 25 male) participated in a study at the DRiVE lab\cite{c33}, which quantified volunteers' DRT to events when a pedestrian laterally intruded into the vehicle's path. Generally, most participants' DRTs were between 0.82 and 2.01 seconds. However, we cannot directly compare our classifier with the human since the two tests are made in totally different environments. So, we only compare the average and range of responding time between our classifier and human to show intuitive results. The range and mean DRT of volunteers in \cite{c33} is $[0.82 s, 2.01 s]$ and 1.46 s. By contrast, our classifier (events that RF $>$ 15 are included) has $[-0.17 s, 0.73 s]$ and 0.19s. From the comparison, we notice that our classifier can respond to a potential collision approximately 1 s prior than the human. If an AV with our approach is moving 50 km/h, then it is able to start braking about 13.89 m earlier than a human driver, not to mention the time advantage of the digital signal over muscle and brake pedal.

It is worth noting that the classifier even recognized the state change before TTEF in 41 sequences. For instance, one should look at the sequence in Fig.~\ref{fig:seq_pose}. We think the reason is that the neural network is able to detect features or preparatory poses, which are almost unperceivable to the human (these appear when pedestrians decide to change crossing states). It supports our opinion that the 2D pose contains enough information to be a strong indicator of the human's crossing state. Moreover, our neural network has also been proved to be powerful to extract and understand features with limited numbers of parameters and shallow layers.

%===============================================================================

%===============================================================================
\section{Conclusions}
\label{sec:conclusion}

In this paper, we extend the C/NC method for AVs to the C/NC/LONG problem and propose a fast shallow neural network classifier for this task. This paper contains extensive validations of our method and reports independent and sequential performance. Promising future work in this area could involve resolving occasional error poses generated by the pose estimator, extracting more robust features from the human pose to improve the classifier's performance, and integrating this with other reliable traffic information to make collision-avoidance systems more robust. Finally, we also hope that future work can compare this type of approaches with human-based systems in well-designed experiments.
%%%%%%%%%%%%%%%%%%%%%%%%%%%%%%%%%%%%%%%%%%%%%%%%%%%%%%%%%%%%%%%%%%%%%%%%%%%%%%%%

\end{document}